\documentclass{article}
\usepackage[utf8]{inputenc}
\usepackage[T1]{fontenc}
\usepackage{microtype}
\usepackage{graphicx}
\usepackage{subfigure}

\usepackage{caption}
\usepackage{booktabs} 

\usepackage[colorinlistoftodos,prependcaption,textsize=tiny]{todonotes}

\usepackage[hyphens]{url}
\usepackage{hyperref}

\usepackage{amsmath,amssymb}
\DeclareMathOperator{\E}{\mathbb{E}}

\usepackage[accepted]{icml2018tadgm}

\icmltitlerunning{Training Strategies for Generative Reversible Networks}

\newcommand{\revised}[1]{#1}
\begin{document}

\twocolumn[
\icmltitle{Training Generative Reversible Networks}



\icmlsetsymbol{equal}{*}

\begin{icmlauthorlist}
\icmlauthor{Robin Tibor Schirrmeister}{tnt,mll}
\icmlauthor{Patryk Chrabąszcz}{mll}
\icmlauthor{Frank Hutter}{mll}
\icmlauthor{Tonio Ball}{tnt}
\end{icmlauthorlist}

\icmlaffiliation{tnt}{Translational Neurotechnology Lab, Medical Center - University of Freiburg, Germany}
\icmlaffiliation{mll}{Machine Learning Lab, University of Freiburg, Germany}

\icmlcorrespondingauthor{Robin Tibor Schirrmeister}{robin.schirrmeister@uniklinik-freiburg.de}

\icmlkeywords{Generative Models, Reversible Networks, Autoencoders}

\vskip 0.3in
]



\printAffiliationsAndNotice{}  

\begin{abstract}
Generative models with an encoding component such as autoencoders currently receive great interest. However, training of autoencoders is typically complicated by the need to train a separate encoder and decoder model that have to be enforced to be reciprocal to each other. To overcome this problem, by-design reversible neural networks (RevNets) had been previously used as generative models either directly optimizing the likelihood of the data under the model or using an adversarial approach on the generated data. Here, we  instead investigate their performance using an adversary on the latent space in the adversarial autoencoder framework. We investigate the generative performance of RevNets on the CelebA dataset, showing that generative RevNets can generate coherent faces with similar quality as Variational Autoencoders. This first attempt to use RevNets inside the adversarial autoencoder framework slightly underperformed relative to recent advanced generative models using an autoencoder component on CelebA, but this gap may diminish with further optimization of the training setup of generative RevNets. In addition to the experiments on CelebA, we show a proof-of-principle experiment on the MNIST dataset suggesting that adversary-free trained RevNets can discover meaningful latent dimensions without pre-specifying the number of dimensions of the latent sampling distribution. In summary, this study shows that RevNets can be employed in different generative training settings.

Source code for this study is at \url{https://github.com/robintibor/generative-reversible}
\end{abstract}

\section{Introduction}

Generative models that include an encoder-decoder architecture have several appealing properties.
For example, they tend to be more stable to train \cite{tolstikhin_wasserstein_2018} and can potentially be used for classification in a semi-supervised fashion \cite{makhzani_adversarial_2016}.
However, a drawback of recent generative models with an encoder-decoder architecture is the requirement to train two separate models, including the need to ensure the encoder and the decoder are reciprocal. While autoencoders with tied weights can at least overcome the problem of training two separate models, recent approaches applying autoencoders to realistic image datasets such as CelebA \cite{liu_deep_2015} use separate decoder and encoder models \cite{donahue_adversarial_2017,dumoulin_adversarially_2016, tolstikhin_wasserstein_2018}.

An interesting alternative to encoder-decoder architectures could be models that are invertible by design.
Recently, invertible-by-design neural networks called reversible neural networks were proposed. In the beginning they were used as generative models \cite{dinh_nice:_2015,dinh_density_2017}, later as classification models with smaller memory requirements \cite{gomez_reversible_2017} and finally to study theoretical assumptions about learning and generalization of deep neural networks \cite{jacobsen_i-revnet:_2018}.
For example, their good classification performance showed that loss of information about the input in later representations of a neural network is not a necessary precondition for good generalization.

\revised{In their application as generative models, reversible networks were trained in two ways. In earlier works, they were trained using the so-called change of variable formula to directly optimize the likelihood of the data under the reversible network model \cite{dinh_nice:_2015,dinh_density_2017}. Later, they were trained  using an adversarial approach on the generated samples \cite{danihelka_comparison_2017,grover_flow-gan:_2018}  same as in generative adversarial networks \cite{goodfellow_generative_2014}.} In this study, we instead investigate their performance when using an adversary in the latent space in an adversarial autoencoder framework \cite{makhzani_adversarial_2016}. In general, the RevNet’s built-in bijectivity could either be an advantage or a disadvantage when optimizing in this framework. 
For example, the bijectivity prevents one from hand-designing the value range for the generated samples as is sometimes done using a sigmoid nonlinearity as the final operation on the decoder output. 

We indeed find it is possible to use RevNets as generative models in the adversarial autoencoder framework, producing samples of comparable quality to variational autoencoders (VAEs) \cite{kingma_auto-encoding_2014} on the CelebA dataset \cite{liu_deep_2015}. Furthermore, in an attempt to exploit the direct correspondence between encodings and inputs in a RevNet, we make a proof of concept for an adversary-free training without a prespecified number of latent dimensions on the MNIST dataset.

\section{Background}
\subsection{Reversible Networks}

\begin{figure}
  \centering
  \includegraphics[width=0.6\linewidth]{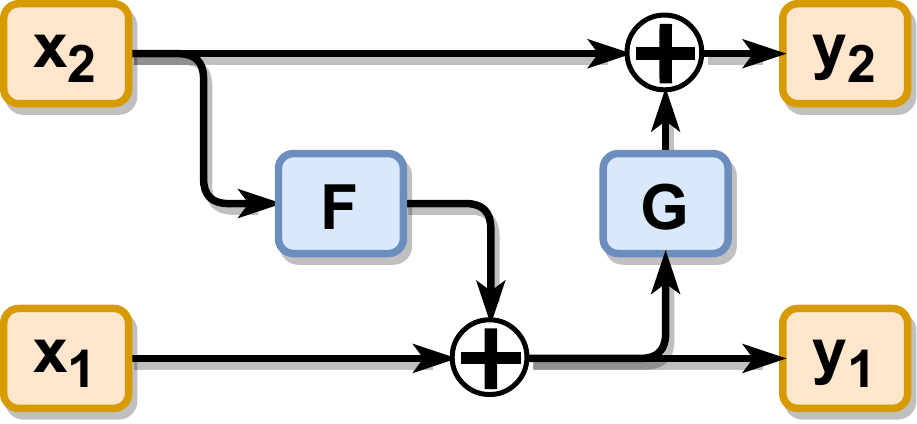} 
  \includegraphics[width=0.6\linewidth]{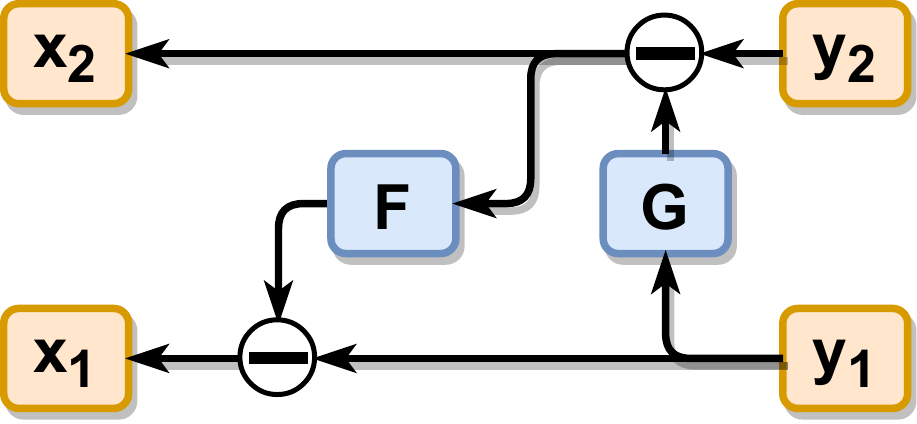} 
  \caption{\textbf{Reversible block.} Functions $F$ and $G$ process inputs $x_1$ and $x_2$ (top), which can be recovered from the outputs $y_1$ and $y_2$ (bottom). See text for details.}
  \label{fig:revnet-functional-block}
\end{figure}

\revised{Reversible networks (RevNets) are neural networks that are invertible by design \cite{dinh_nice:_2015,gomez_reversible_2017,jacobsen_i-revnet:_2018} through the use of invertible blocks.}
The basic invertible block is defined for an input $x$, split into disjoint parts $x_1$,$x_2$ and two functions F and G that have the same output as input size as follows (also see Figure \ref{fig:revnet-functional-block}):

\begin{equation}\label{eq:reversible-forward}
  \begin{aligned}
  y_1 = F(x_2) + x_1 \\
  y_2 = G(y_1) + x_2
  \end{aligned}
\end{equation}

Inputs $x_1$ and $x_2$ can be inverted from the outputs $y_1$ and $y_2$ as follows:

\begin{equation}\label{eq:reversible-backward}
  \begin{aligned}
  x_2 = y_2 - G(y_1) \\
  x_1 = y_1 - F(x_2)
  \end{aligned}
\end{equation}

$F$ and $G$ will typically be a sequence of convolutional or other neural network layers.
The splitting of input x into disjoint inputs $x_1$, $x_2$ is often implemented along the channel dimension of the network. 

\begin{figure}
  \begin{center}
  \includegraphics[width=\linewidth]{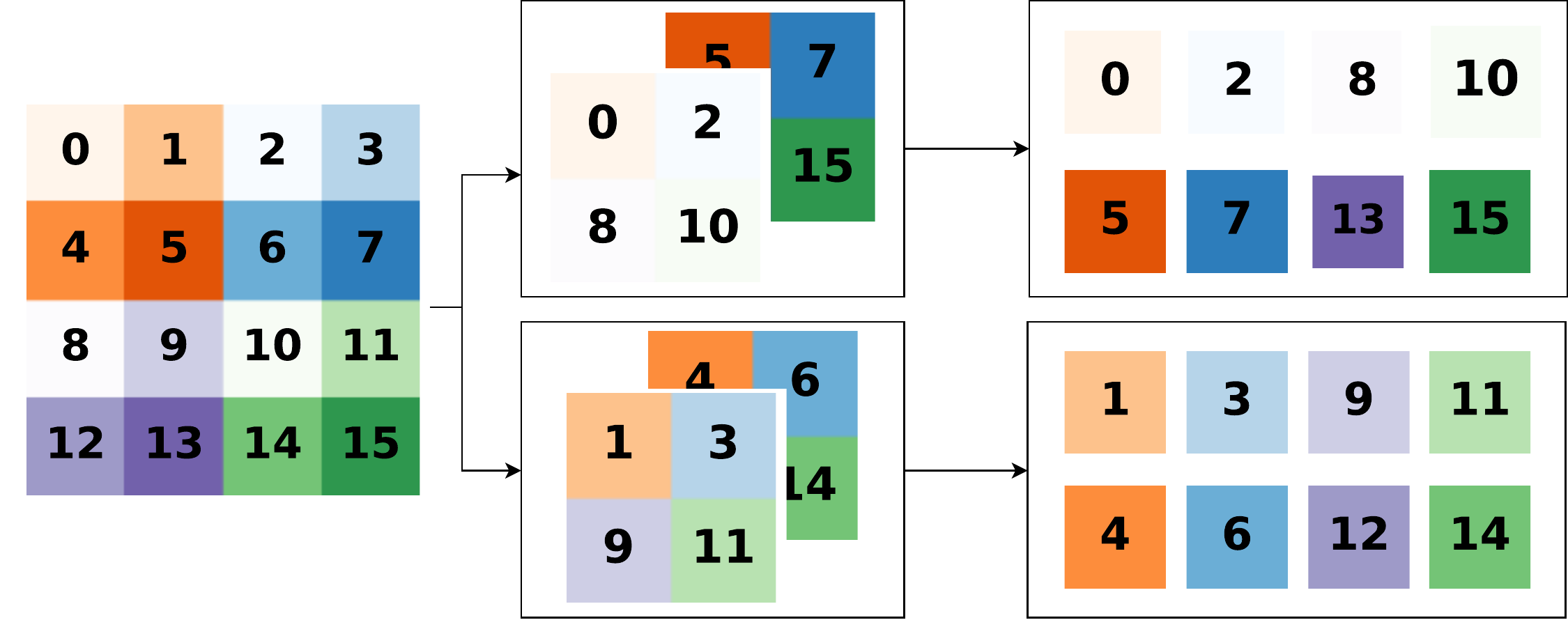}
  \end{center}
    \caption{\textbf{Subsampling step}. 
    Our subsampling operation applied twice to a 4x4 input. Upper and lower rectangles on middle and right represent the different streams inside the reversible net (i.e., $x_1$ and $x_2$). On the right, individual squares represent individual channels, so each channel has a single value at the end. Note that at the end, both streams have access to pixels that cover the entire 4x4 input.}
    \label{fig:subsampling}
\end{figure}

\begin{figure}
  \begin{center}
  \includegraphics[width=\linewidth]{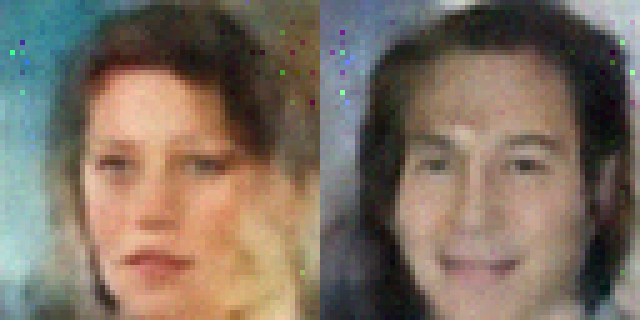}
  \end{center}
    \caption{\textbf{Artifacts in earlier phases of the training}. 
    Generated samples on CelebA of an uncoverged RevNet with green and purple pixel artifacts. Artifacts are caused by some latent dimensions still strongly influencing the input dimensions they correspond to as explained in the text.}
    \label{fig:artifacts}
\end{figure}

One important addition to the invertible network architecture are invertible subsampling blocks that were introduced
to make the RevNets end-to-end invertible \cite{dinh_nice:_2015,jacobsen_i-revnet:_2018}.
Invertible subsampling is possible by shifting spatial dimensions into the channel dimensions.
Basically, for a 2x2 subsampling, 4 translated spatial checkerboard patterns of the input are moved into four different channels as seen in Figure \ref{fig:subsampling}.
Our subsampling operation is a slightly modified version of the operation proposed in  earlier work \cite{dinh_nice:_2015,jacobsen_i-revnet:_2018} that ensures that the final $y_1$ and $y_2$ correspond to checkerboard patterns covering the entire input image as indicated in Figure \ref{fig:subsampling}.
This was motivated by our observation that early on in the training, the values of the RevNet encodings are still strongly influenced by the values at the input positions they correspond to, as shown in Figure \ref{fig:artifacts}. Therefore, both F and G seeing inputs that cover the entire image might make it easier for F and G to correctly predict what will be added to their output, easing the generative training.

\subsection{Adversarial Autoencoders}
In the adversarial autoencoder framework, the encoder and decoder are trained together to minimize a reconstruction loss on the decoded inputs and an adversarial loss on the encodings. 
For the reconstruction loss, the encoder and decoder are optimized to minimize a reconstruction error of the inputs 
$\E_{x_\sim P_{x}}c(x, G(E(x)))$, where $x$ is an input, $P_x$ the input distribution, $G$ is the decoder (generator), $E$ is the encoder, and $c$ is a reconstruction loss such as L1-loss or L2-loss.

Since RevNets are invertible by construction, we propose to use a single RevNet to instantiate both the encoder and the decoder, leading to a reconstruction loss of zero by design, regardless of the weights used in the RevNet. In practice, we aim for a lower-dimensional latent space that still yields good reconstructions. In order to obtain this, in the reconstruction training phase, we clip the encodings produced by the RevNet according to a prior distribution that sets most encoding dimensions to zero; we then invert the clipped encodings through the RevNet and optimize the L1 reconstruction loss between the original inputs and the inverted clipped encodings, which has been reported to work better than L2 in natural image settings \cite{isola_image--image_2017,ulyanov_adversarial_2017}.
We also penalize the L2-distance of the encodings and the clipped encodings as we found this to greatly stabilize this training phase.\footnote{In practice, since we wanted to keep the option of using a uniform distribution in +-2 as the latent sampling distribution, we also clipped the nonzero dimensions to +-2 for both the L1 and L2 loss, but we do not expect this to strongly influence the results.}

While in principle the reconstruction phase is not even necessary for reversible networks, we still found it useful as a first phase to allow the network to generate a useful arrangement of the inputs in the encoding space before optimization of the adversarial loss. 
We still apply this reconstruction loss in the next phase of the training where we include an adversarial loss.

For the adversarial loss, a discriminator network is trained to distinguish the distribution of the encoder outputs from a prior distribution. The encoder tries to fool the discriminator by making the encoder outputs indistinguishable from samples of the prior distribution. The adversarial game can be setup with a variety of loss functions; we choose the adversarial hinge loss as advocated for the use in Generative Adversarial Networks (GANs) \cite{goodfellow_generative_2014} in \citet{lim_geometric_2017}:

\begin{equation}\label{eq:hinge-loss-adversarial}
  \begin{aligned}
   L_D =& -\E_{z\sim P_{prior}}[\min(0,-1+D(z)]  \\ 
   & - \E_{x\sim P_{X}}[\min(0,-1-D(E(x)))] \\
   L_E =& -\E_{x\sim P_{X}}[D(E(x))]
  \end{aligned}
\end{equation}

with $L_D$ and $L_E$ the losses for the discriminator and the encoder, in our case the RevNet, respectively. 
In our setting, we only apply the discriminator on the nonzero dimensions of the prior distribution, while penalizing the remaining dimensions through the L1 and L2 loss on the clipped encodings as explained before. This should greatly simplify the adversarial training as it makes the problem substantially lower dimensional (e.g., 64 dimensions vs. 12288 dimensions in the case of a 64-dimensional prior and 64x64 RGB images, which will be our setting on the CelebA \cite{liu_deep_2015} dataset as explained in the experiments section).

We also make use of the recently proposed spectral normalization for the discriminator \cite{miyato_spectral_2018}. 
Spectral normalization normalizes the spectral norm $\sigma(W)$ of any weight matrix $W$ of the discriminator to unit spectral norm:

\begin{equation}\label{eq:spectral norm}
  \begin{aligned}
   W_{SN}(W) =& \ W/\sigma(W)\\ 
   \sigma(W) =& \ \max_{h:h\neq0}\frac{||Wh||_2}{||h||_2}
  \end{aligned}
\end{equation}

where $\sigma(W)$ is also equivalent to the largest singular value of $W$.
Spectral normalization was designed to regularize the Lipschitz norm of the discriminator network to stabilize the training \cite{miyato_spectral_2018}.
In practice, spectral normalization can be computed efficiently using the power iteration method, only using a single iteration per forward pass; we defer to \citet{miyato_spectral_2018} for details.

\subsection{Optimal Transport}
Optimal transport distances measures the distance between two distributions as the distance needed to morph one distribution into the other.
This can be visualized as the transport of sand when imagining both distributions to be piles of sand \cite{peyre_computational_2018}. Formally, it is defined for two distributions $P_x, P_y$ as:

\begin{equation}\label{eq:transport-distance}
  \begin{aligned}
   OT(P_x, P_y)=\inf_{\Gamma \in \mathcal{P}(x\sim P_x, y \sim P_y)}=\E_{(x,y)\sim \Gamma} [c(x,y)],
  \end{aligned}
\end{equation}

where $c(x,y)$ is a user-defined cost/distance function, and $\Gamma$ is a coupling distribution whose probabilities specifies how much probability mass is moved from each point $x\in X$ to each point $y \in Y$. To ensure that this coupling correctly distributes all the mass from one distribution to the other, it must come from the set $\mathcal{P}(x\sim P_x, y \sim P_y)$ of all joint distributions of $(x,y)$ with marginals $P_x$ and $P_y$, respectively. For two empirical distributions with the same number of samples, it is equivalent to the pairing that minimizes the average distance between the pairs.

Optimal transport distances have recently seen an increasing usage and interest in the field of generative models, especially due to their ability to compare distributions with disjoint support. As such, they have been used in different ways to train GANs \cite{arjovsky_wasserstein_2017,salimans_improving_2018}.
For a more thorough overview over optimal transport and its applications, we highly recommend \citet{peyre_computational_2018}.
In this study, we make more direct use of optimal transport distances in our experiments on the MNIST dataset \cite{lecun_gradient-based_1998}.

Finally, we note that theoretical analysis using optimal transport distances has recently generalized the adversarial autoencoder framework into the Wasserstein Autoencoder framework \cite{tolstikhin_wasserstein_2018}. This analysis showed that any method that matches the latent sampling distribution and the encoding distribution of the real inputs can minimize an arbitrary optimal transport distance (the chosen reconstruction loss) between the distribution of generated inputs and the real input distribution. More precisely, for a given decoder, a given latent sampling distribution and a given distance, the optimal transport distance is equivalent to the minimum expected encoder-decoder reconstruction distance over all such encoders whose encoding distribution of the real inputs is identical to the latent sampling distribution. We defer to \citet{tolstikhin_wasserstein_2018} for more details. 

For our RevNets, inverting unclipped encodings of a RevNet should result in the exact same inputs that produced the encodings.
Therefore, the distribution of generated samples would be identical to the real input distribution if the latent sampling distribution would match the encoding distribution of the real inputs produced by the RevNets exactly. Nevertheless, as the encodings never exactly match the imposed prior distribution, it remains important that the encoding distances remain meaningful throughout the training, which we found to be much more so when using the initial reconstruction  phase described earlier.

\subsection{Fréchet Inception Distance}

The Fréchet Inception Distance (FID) has been proposed as a measure for evaluating the quality of generated samples for a specific dataset \cite{heusel_gans_2017}. It is the optimal L2-transport distance between features of the Imagenet-pretrained Inception network computed on the given dataset and computed on a set of generated samples, under the assumption that both feature distributions follow a Gaussian distribution. The Gaussianity assumption makes it possible to compute the optimal transport distance directly from the mean and covariance matrices. The FID has been advocated as the measure best correlated with human notions of sample quality of all automatically computable measures that have been proposed so far \cite{heusel_gans_2017, lucic_are_2017}, although recently alternatives overcoming the assumption of Gaussianality have been proposed \cite{binkowski_demystifying_2018}.

\section{Experiments}
\subsection{CelebA}
\begin{figure}
  \begin{center}
  \includegraphics[width=0.2\linewidth]{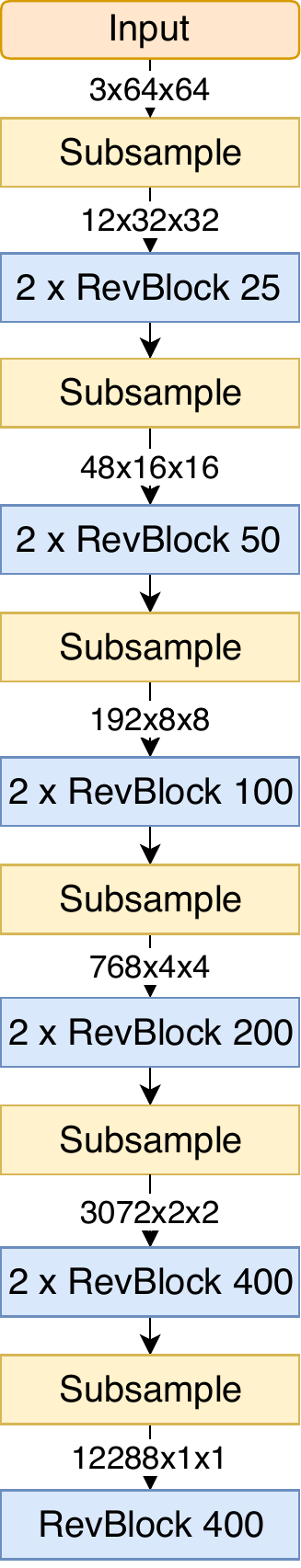}
  \hspace{1cm}
   \includegraphics[width=0.3\linewidth]{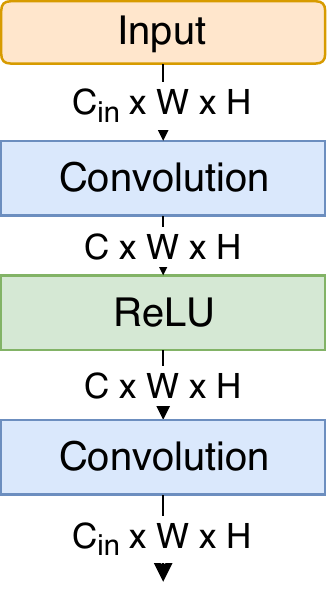}
  \end{center}
    \caption{\textbf{RevNet Architecture and F/G functions}. On the left, RevNet architecture for use on CelebA. On the right, our F/G function inside the reversible blocks. 
    The RevNet functions both as the encoder (from top to bottom) and through its inverse as the decoder (from bottom to top).
    Numbers after each RevBlock indicate intermediate number of channels. For the F/G function, $C_{\text{in}}$ indicates the number of input channels for the function and $C$ indicates the aforementioned intermediate number of convolutional filters/channels.}
    \label{fig:revnet-architecture}
\end{figure}

We run our our generative reversible network on the CelebA dataset \cite{liu_deep_2015}, a widely used dataset to evaluate autoencoders.
We crop and downsample the images to 64x64 pixels as is common practice, using the same code as in \citet{tolstikhin_wasserstein_2018}\footnote{See code here: \url{https://github.com/tolstikhin/wae/blob/a1fdf24066b83665feffbcf18298cd605658e33d/datahandler.py##L188-L208}}.
Our RevNet architecture uses 11 reversible function blocks and 6 reversible subsampling steps with 60 million parameters and is shown in Figure \ref{fig:revnet-architecture}.

For the discriminator, we use a fully connected network with 2 hidden layers with 400 and 800 units each.
The first layer uses concatenated ReLUs ($x\rightarrow (max(x,0), max(-x,0)$) \cite{shang_understanding_2016}  and the second layer regular ReLUs \cite{glorot_deep_2011} as nonlinearities.
We chose concatenated ReLUs in the first layer as we observed in preliminary experiments that they help the discriminator produce more useful gradients when the encodings are too concentrated around the mean of the prior distribution. 
We apply spectral normalization on the discriminator using 1 power iteration per forward pass as described in \citet{miyato_spectral_2018}. 

The prior distribution is a 64-dimensional standard-normal distribution.
The 64 dimensions are the output dimensions with the highest standard deviations of the outputs for the untrained RevNet on the dataset.
For the optimization, we follow \citet{heusel_gans_2017} in employing different learning rates for the generative RevNet and the discriminator, using Adam with $\alpha=1e-4$ and $\alpha=4e-4$, respectively ($\beta_1=0$ and $\beta_2=0.9$ for both). These settings were chosen identical to a fairly recent successful GAN setting \cite{zhang_self-attention_2018}. Code for reproducing these experiments will be released upon publication.

\subsubsection {Results}
\begin{figure}[tbp]
  \begin{center}
  \includegraphics[width=\linewidth]{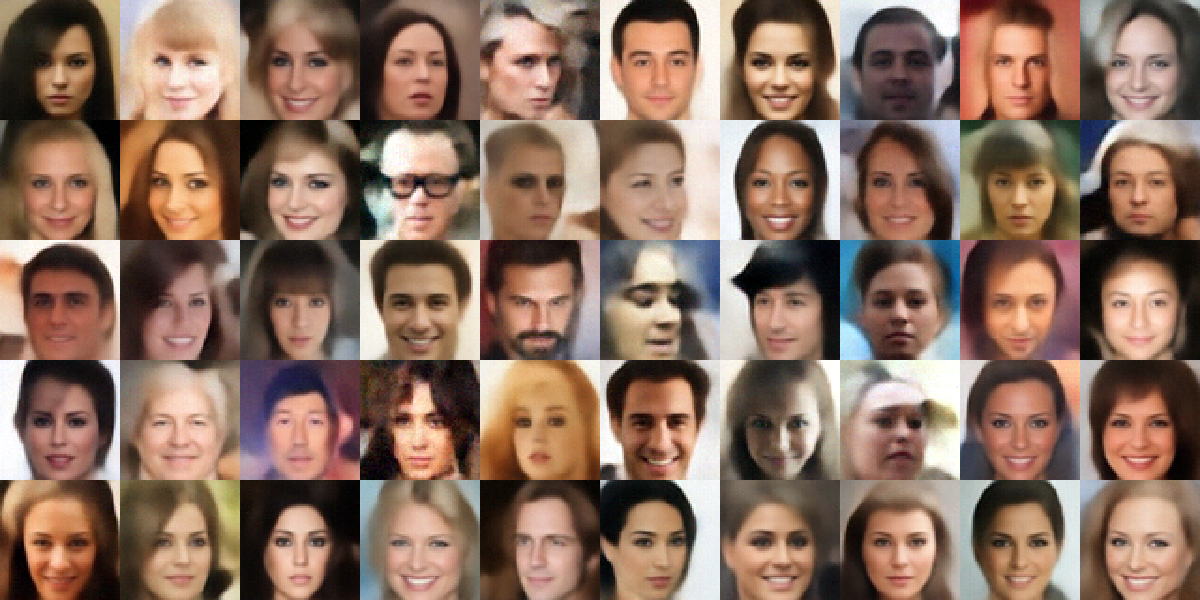}
  \end{center}
    \caption{\textbf{Generated samples on CelebA.} Samples are for the most part globally coherent, lacking some details.}
    \label{fig:celeba-samples}
\end{figure}

\begin{figure}[tbp]
  \begin{center}
  \includegraphics[width=\linewidth]{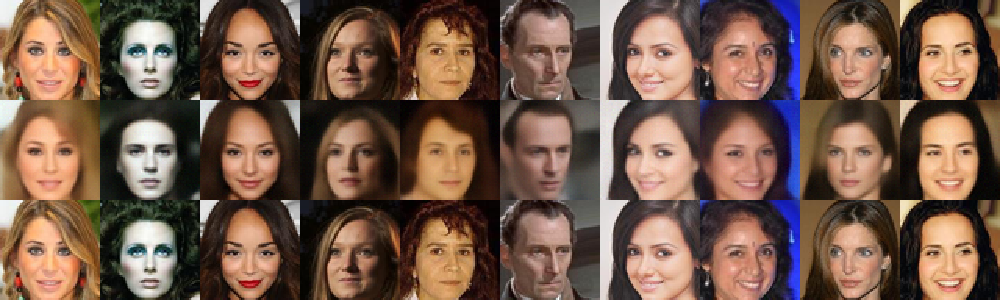}
  \end{center}
    \caption{\textbf{Reconstructions on CelebA.} Top row: original, middle row: reconstruction from latent space restricted to the prior distribution, bottom row: reconstruction from full latent space. Reconstructions from restricted latent space are somewhat blurry and lose detail, reconstructions from full latent space show that numerical errors do not lead to visible image changes.}
    \label{fig:celeba-reconstructions}
\end{figure}

\begin{figure}[tbp]
  \begin{center}
  \includegraphics[width=\linewidth]{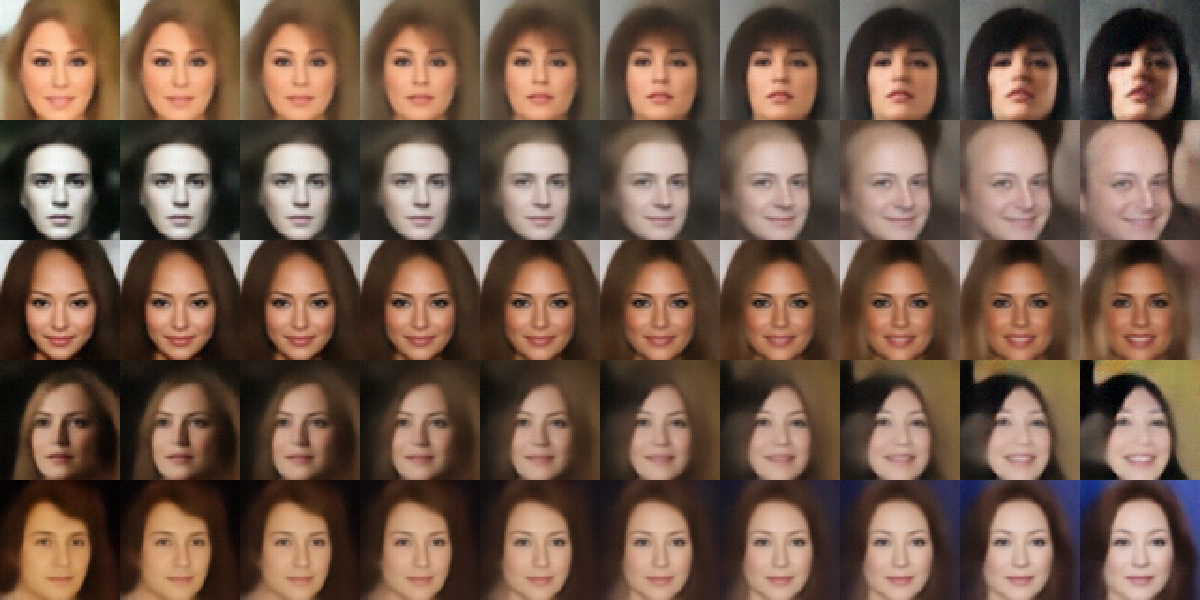}
  \end{center}
    \caption{\textbf{Interpolations on CelebA in restricted latent space.} Images obtained by interpolating between encodings of two inputs in the encoding space restricted to the nonzero dimensions of the latent sampling distribution. Intermediate images clearly resemble human faces.}
    \label{fig:celeba-interpolations-restricted}
\end{figure}

\begin{figure}[tbp]
  \begin{center}
  \includegraphics[width=\linewidth]{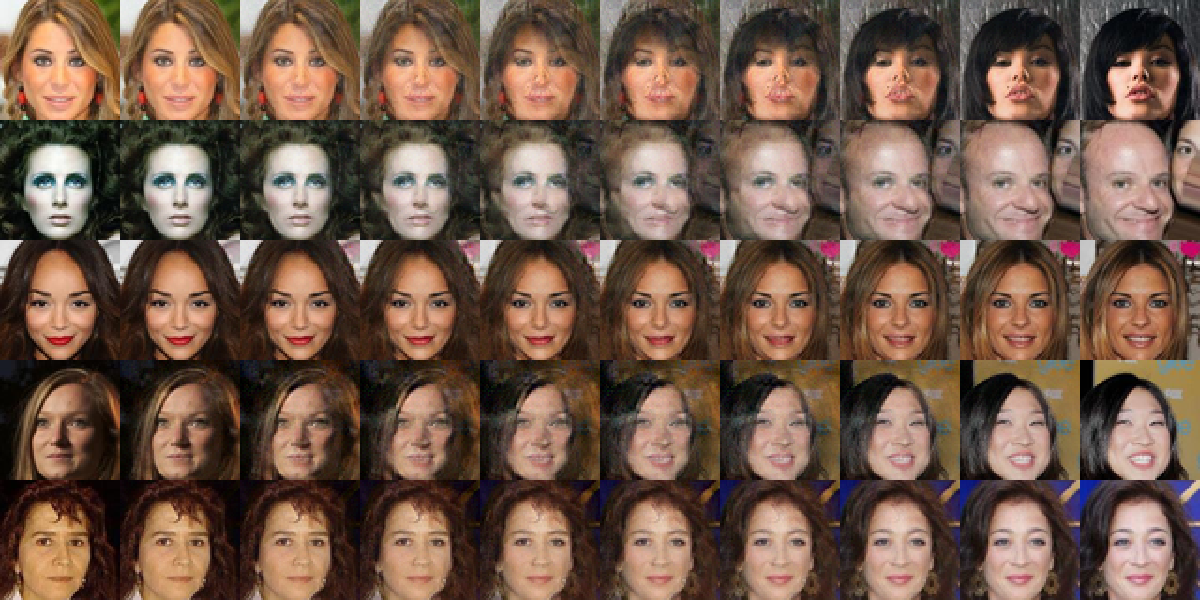}
  \end{center}
    \caption{\textbf{Interpolations on CelebA in full latent space.} Intermediate images have more details, however show clear unnatural artifact patterns.}
    \label{fig:celeba-interpolations-full}
\end{figure}

\begin{figure}[tbp]
  \begin{center}
  \includegraphics[width=\linewidth]{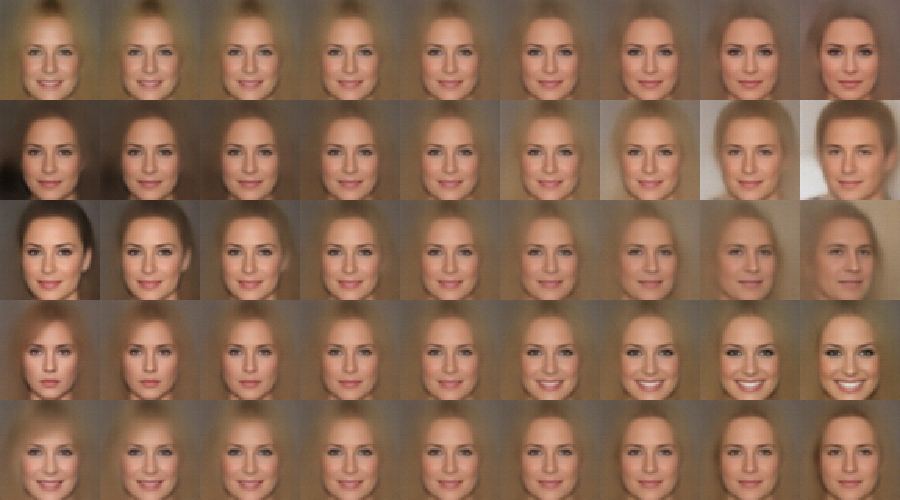}
  \end{center}
    \caption{\textbf{Samples when varying five randomly chosen dimensions of the latent prior.} Top to bottom: Different latent dimensions. Left to right: Varying the corresponding latent dimension from -3 to +3 standard deviations around the mean. Different latent dimensions seem to encode different combinations of attributes such as smiling vs nonsmiling, hair color, gender, background color, etc.}
    \label{fig:fig:celeba-varying-latent}
\end{figure}

Our generative reversible network generates globally coherent faces as seen in Figure \ref{fig:celeba-samples}.
The generated faces are fairly blurry, which is also reflected in an FID score close to those reported for VAEs and higher than for other autoencoders in an adversarial framework (see Table \ref{tab:fid}).
Reconstructions from the restricted latent space again show that the RevNet preserves some global attributes while losing detail (Figure \ref{fig:celeba-reconstructions}).
Reconstructions from the unrestricted outputs show that the RevNet does not suffer from any numerical instabilities that can be visually perceived from the reconstructions (Figure \ref{fig:celeba-reconstructions}). 
Numerical analysis of the reconstruction losses confirms this with a mean L1 error of $9e-7$ on the entire CelebA dataset for our trained RevNet. 
Interpolations in latent space show coherent interpolated faces when staying in the latent space restricted to the nonzero dimensions of the prior (Figure \ref{fig:celeba-interpolations-restricted}).
Interpolations in the full latent space, while having more detail, also show unrealistic artifacts in some cases (Figure \ref{fig:celeba-interpolations-full}).
Samples generated by varying the latent space in 5 dimensions of the prior latent distribution show that the latent dimensions seem to encode combinations of semantically meaningful attributes such as smiling vs. nonsmiling, hair color, background and gender (Figure \ref{fig:fig:celeba-varying-latent}).

Finally, we observe that the training is very stable, we rerun the experiment 4 times using the same model pretrained in the reconstruction phase but varying the order of examples and the seeds for initializing the adversary parameters. Due to time constraints, we were not yet able to rerun the reconstruction phase with different seeds, but based on preliminary experiments we expect similar results in that case as well.

\begin{table}[t]
\caption{\textbf{Fréchet Inception distance on CelebA.} Estimated from 10000 samples to be consistent to \cite{tolstikhin_wasserstein_2018} (lower is better).}
\label{tab:fid}
\vskip 0.15in
\begin{center}
\begin{small}
\begin{sc}
\begin{tabular}{lcccr}
\toprule
Model & FID \\
\midrule
Variational \\ Autoencoder    & 63  \\
WAE-MMD & 55  \\
WAE-GAN & 42  \\
RevNet    & 65\\
\bottomrule
\end{tabular}
\end{sc}
\end{small}
\end{center}
\vskip -0.2in
\end{table}

\begin{figure}[tbp]
  \begin{center}
  \includegraphics[width=0.8\linewidth]{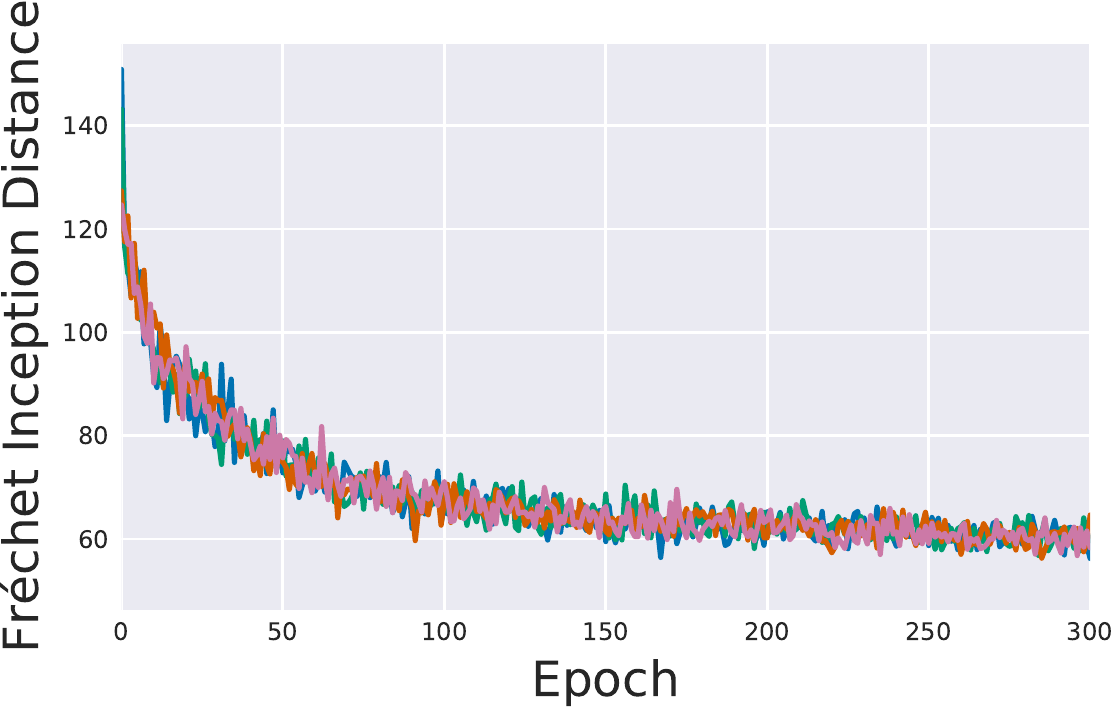}
  \end{center}
    \caption{\textbf{Learning curves for the Fréchet Inception Distance.} Epoch refers to one training epoch passing over the whole dataset. Colors indicate different runs using a different order of examples and different seeds for initializing the adversary. Curves are very similar, showing very stable training. Note that these FID scores are computed using the PyTorch Inception model and differ from the corresponding scores computed with Tensorflow (hence the discrepancy to Table \ref{tab:fid}).}
    \label{fig:celeba-fid-curves}
\end{figure}

 \vspace*{2cm}

\subsection{MNIST}

In a second experiment, we attempted to answer two questions: First, can generative reversible networks be trained using optimal transport without an adversary? The question of adversary-free training or alternatively, training with a adversary limited to computing an adversarial kernel function, continues to attract considerable interest due to the often difficult training dynamics of generative adversarial networks \cite{binkowski_demystifying_2018,tolstikhin_wasserstein_2018, rubenstein_latent_2018}. Second, is it possible to avoid prespecifying the latent dimensionality? This question is interesting as using a too large latent dimensionality might make the matching impossible \cite{makhzani_adversarial_2016,tolstikhin_wasserstein_2018,rubenstein_latent_2018} and using a too small latent dimensionality might make the network unable to model some variation in the generated samples, which it could otherwise retain (see \citet{rubenstein_latent_2018} for a more thorough discussion of these effects).

We chose the MNIST dataset as this experiment should mainly serve as a proof of principle, and not to judge the quality of this approach compared to more established approaches of optimizing generative models. To this end, we considered that a simpler dataset, such as MNIST, with less factors of variation, could yield more helpful insights for a first attempt.

Concretely, we train the RevNet to match class-conditional latent distributions on the outputs, while we optimize the parameters of these distributions at the same time as follows. We first define the class-conditional latent distributions as uncorrelated Gaussian distributions and set the means and standard deviations to the corresponding means and standard deviations of the encodings of the untrained RevNet. Then, for each minibatch, we compute the optimal transport distance for the encodings of that minibatch and a same-size sample from the latent distribution using Euclidean distances as the cost function and solving the transport problem exactly using the algorithm from \citet{bonneel_displacement_2011} \footnote{We use the code from the Python Optimal Transport library, \url{https://github.com/rflamary/POT/blob/81b2796226f3abde29fc024752728444da77509a/ot/lp/__init__.py##L19}}. The optimal transport distance is then used as a loss for both the RevNet and the means and standard deviations of the class-conditional latent distributions. While the optimal transport distance is known to have biased gradients \cite{bellemare_cramer_2017,salimans_improving_2018}, we still find it to work well enough on MNIST for reasonable per-class batch sizes ($<1000$). 
Besides ensuring a low optimal transport distance of the encodings and the sampling distributions, we must prevent the RevNet from ``hiding'' information in encoding dimensions with small standard deviations for these transport distances in encoding space to remain meaningful and to keep the training stable. For that, we propose a simple perturbation loss, that penalizes the reconstruction loss after applying a small perturbation sampled from a Gaussian distribution on the encodings. Concretely we penalize:

\begin{equation}\label{eq:perturbation-loss}
  \begin{aligned}
   L_{perturb} =& \E_{x\sim P_{X}}[c(x,R^{-1}(R(x)+\epsilon)) \\
   \epsilon \sim& \mathcal{N}(0,0.01) 
  \end{aligned}
\end{equation}

where $R$ and $R^{-1}$ are the forward and inverse functions of the RevNet, respectively. The perturbation loss should also prevent the RevNet and the latent distributions from shrinking their standard deviations too much, which would otherwise cause a very unstable training which we have also observed in practice.

\begin{figure}
    \centering
      \includegraphics[width=0.9\linewidth]{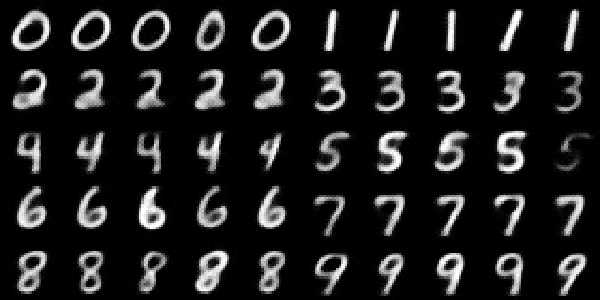} 
     \caption{\textbf{Samples on MNIST.} Samples show realistic digits, are somewhat blurry and lack some diversity.}
     \label{fig:mnist-samples}
\end{figure}

\begin{figure}
    \centering
        \includegraphics[width=0.9\linewidth]{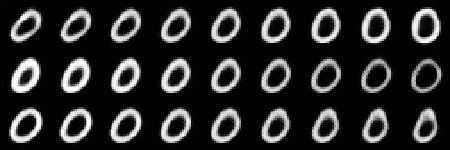} 
        \includegraphics[width=0.9\linewidth]{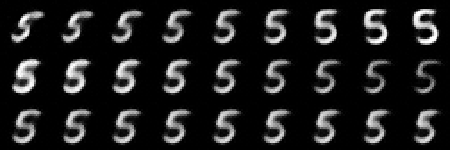} 
        \includegraphics[width=0.9\linewidth]{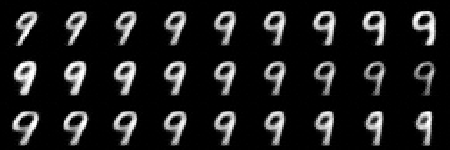}
     \caption{\textbf{Samples when varying three dimensions of the latent prior.} Varying the three latent dimensions with the largest standard deviations of the averaged per-class standard deviations. The three dimensions seem to roughly correspond to tilt, thickness and size respectively.}
     \label{fig:mnist-varying-latent} 
\end{figure}

\subsubsection{Results}

The RevNet ends up using only 3-4 dimensions per class to encode the digits, with several of the dimensions shared between the classes. While the samples are somewhat blurry and lack diversity (see Figure \ref{fig:mnist-samples}), interestingly some of the used dimensions with the largest standard deviations encode semantically identical features for the different digits as shown in Figure \ref{fig:mnist-varying-latent}.
This indicates the RevNet has learned an encoding that keeps class-independent dimensions such as thickness or tilt in the same encoding dimensions despite having the freedom to use completely different dimensions for the different classes.

\section{Discussion}
Overall, we have shown for the first time that reversible neural networks can be used inside the adversarial autoencoder framework, yielding globally coherent generated faces on CelebA.
While they still underperform relative to recent advanced generative autoencoder models on that dataset according to the Fréchet Inception Distance, the performance gap might be due to hyperparameters or architecture design choices, which have not been explored for RevNets prior to this work and are known to strongly affect generative model results \cite{lucic_are_2017}.
Closing the performance gap through automated search for architectures and hyperparameters could therefore be an interesting next step. This could also include other forms of matching the distributions such as maximum mean discrepancy \cite{gretton_kernel_2012, li_mmd_2017,tolstikhin_wasserstein_2018} or sliced Wasserstein distances \cite{kolouri_sliced-wasserstein_2018}.

\revised{Furthermore, the previous maximum-likelihood and input-adversarial methods used to train invertible networks in a generative setting \cite{dinh_nice:_2015,dinh_density_2017, danihelka_comparison_2017,grover_flow-gan:_2018} could be more directly compared to the adversarial autoencoder method from this study. The generated samples in the maximum-likelihood approach on CelebA in \citet{dinh_density_2017}  feature more details, but also more unnatural artifacts. Attributing these differences to the training procedure or the model architecture could meaningfully extend prior work comparing maximum-likelihood and input-adversarial training of generative RevNets \cite{danihelka_comparison_2017}. For the input-adversarial approach, one could also combine it with our proposal to use only a subset of the full latent sampling dimensionality. For this combination, it might be insightful to study the resulting encodings of the real inputs, especially in terms of what is modelled in the encoding space outside of the used sampling distribution, similar to our reconstructions from restricted and full latent space. Finally, one could compare the performance of RevNets n the adversarial autoencoder framework to approaches that use more traditional non-invertible autoencoders \cite{donahue_adversarial_2017,dumoulin_adversarially_2016,ulyanov_adversarial_2017}}

\revised{Later works on generative invertible networks used a hierarchical ordering of the latent sampling dimensions (see \citet{dinh_density_2017} for details). This might be worth exploring further. First, one might study this idea in combination with the adversarial autoencoder framework employed in this study. Second, the  model architecture and hierarchical latent dimension ordering to enable high-quality generative modelling could be further optimized. Third, one might try to combine this idea with the progressive training of generative models as in \citet{karras_progressive_2018}.}

Our experiment on MNIST indicates that for simple datasets an adversary-free approach that does not need a prespecified latent dimensionality can result in  meaningful encoding dimensions. This might be interesting for other works investigating the effect of latent dimensionality and intrinsic dimensionality on generative models \cite{rubenstein_latent_2018}. However, even for MNIST, the results are somewhat underwhelming with regards to the diversity of the generated samples. Still, we hope our results inspire further investigations on how to properly achieve the goals of having a meaningful encoding dimension, a small distance between encodings of the real inputs and the sampling distribution and realistic generated samples.

Additionally, the excellent performance of reversible networks on supervised classification tasks \cite{jacobsen_i-revnet:_2018} makes it attractive to investigate their use in semi-supervised classification settings where adversarial autoencoders have already shown good results \cite{makhzani_adversarial_2016}.

\section*{Acknowledgements}

This work was supported by the BrainLinks-BrainTools Cluster of Excellence (DFG grant EXC 1086) and by the Federal 
Ministry of Education and Research (BMBF, grant Motor-BIC 13GW0053D).

\bibliography{GenerativeReversible}

\begin{thebibliography}{34}
\providecommand{\natexlab}[1]{#1}
\providecommand{\url}[1]{\texttt{#1}}
\expandafter\ifx\csname urlstyle\endcsname\relax
  \providecommand{\doi}[1]{doi: #1}\else
  \providecommand{\doi}{doi: \begingroup \urlstyle{rm}\Url}\fi

\bibitem[Arjovsky et~al.(2017)Arjovsky, Chintala, and
  Bottou]{arjovsky_wasserstein_2017}
Arjovsky, M., Chintala, S., and Bottou, L.
\newblock Wasserstein {GAN}.
\newblock \emph{arXiv:1701.07875 [cs, stat]}, January 2017.
\newblock URL \url{http://arxiv.org/abs/1701.07875}.
\newblock arXiv: 1701.07875.

\bibitem[Bellemare et~al.(2017)Bellemare, Danihelka, Dabney, Mohamed,
  Lakshminarayanan, Hoyer, and Munos]{bellemare_cramer_2017}
Bellemare, M.~G., Danihelka, I., Dabney, W., Mohamed, S., Lakshminarayanan, B.,
  Hoyer, S., and Munos, R.
\newblock The {Cramer} {Distance} as a {Solution} to {Biased} {Wasserstein}
  {Gradients}.
\newblock \emph{arXiv:1705.10743 [cs, stat]}, May 2017.
\newblock URL \url{http://arxiv.org/abs/1705.10743}.
\newblock arXiv: 1705.10743.

\bibitem[Bińkowski et~al.(2018)Bińkowski, Sutherland, Arbel, and
  Gretton]{binkowski_demystifying_2018}
Bińkowski, M., Sutherland, D.~J., Arbel, M., and Gretton, A.
\newblock Demystifying {MMD} {GANs}.
\newblock In \emph{International {Conference} on {Learning} {Representations}
  ({ICLR})}, 2018.
\newblock URL \url{https://openreview.net/forum?id=r1lUOzWCW}.

\bibitem[Bonneel et~al.(2011)Bonneel, van~de Panne, Paris, and
  Heidrich]{bonneel_displacement_2011}
Bonneel, N., van~de Panne, M., Paris, S., and Heidrich, W.
\newblock Displacement {Interpolation} {Using} {Lagrangian} {Mass} {Transport}.
\newblock In \emph{Proceedings of the 2011 {SIGGRAPH} {Asia} {Conference}},
  {SA} '11, pp.\  158:1--158:12, New York, NY, USA, 2011. ACM.
\newblock ISBN 978-1-4503-0807-6.
\newblock \doi{10.1145/2024156.2024192}.
\newblock URL \url{http://doi.acm.org/10.1145/2024156.2024192}.

\bibitem[Danihelka et~al.(2017)Danihelka, Lakshminarayanan, Uria, Wierstra, and
  Dayan]{danihelka_comparison_2017}
Danihelka, I., Lakshminarayanan, B., Uria, B., Wierstra, D., and Dayan, P.
\newblock Comparison of {Maximum} {Likelihood} and {GAN}-based training of
  {Real} {NVPs}.
\newblock \emph{arXiv:1705.05263 [cs]}, May 2017.
\newblock URL \url{http://arxiv.org/abs/1705.05263}.
\newblock arXiv: 1705.05263.

\bibitem[Dinh et~al.(2015)Dinh, Krueger, and Bengio]{dinh_nice:_2015}
Dinh, L., Krueger, D., and Bengio, Y.
\newblock {NICE}: {Non}-linear {Independent} {Components} {Estimation}.
\newblock In \emph{International {Conference} on {Learning} {Representations}
  ({ICLR})}, 2015.
\newblock URL \url{http://arxiv.org/abs/1410.8516}.
\newblock arXiv: 1410.8516.

\bibitem[Dinh et~al.(2017)Dinh, Sohl-Dickstein, and Bengio]{dinh_density_2017}
Dinh, L., Sohl-Dickstein, J., and Bengio, S.
\newblock Density estimation using {Real} {NVP}.
\newblock In \emph{International {Conference} on {Learning} {Representations}
  ({ICLR})}, 2017.
\newblock URL \url{https://openreview.net/forum?id=SyPNSAW5}.

\bibitem[Donahue et~al.(2017)Donahue, Krähenbühl, and
  Darrell]{donahue_adversarial_2017}
Donahue, J., Krähenbühl, P., and Darrell, T.
\newblock Adversarial {Feature} {Learning}.
\newblock In \emph{International {Conference} on {Learning} {Representations}
  ({ICLR})}, 2017.
\newblock URL \url{http://arxiv.org/abs/1605.09782}.
\newblock arXiv: 1605.09782.

\bibitem[Dumoulin et~al.(2016)Dumoulin, Belghazi, Poole, Mastropietro, Lamb,
  Arjovsky, and Courville]{dumoulin_adversarially_2016}
Dumoulin, V., Belghazi, I., Poole, B., Mastropietro, O., Lamb, A., Arjovsky,
  M., and Courville, A.
\newblock Adversarially {Learned} {Inference}.
\newblock \emph{arXiv:1606.00704 [cs, stat]}, June 2016.
\newblock URL \url{http://arxiv.org/abs/1606.00704}.
\newblock arXiv: 1606.00704.

\bibitem[Glorot et~al.(2011)Glorot, Bordes, and Bengio]{glorot_deep_2011}
Glorot, X., Bordes, A., and Bengio, Y.
\newblock Deep sparse rectifier neural networks.
\newblock In \emph{Proceedings of the {Fourteenth} {International} {Conference}
  on {Artificial} {Intelligence} and {Statistics}}, pp.\  315--323, 2011.

\bibitem[Gomez et~al.(2017)Gomez, Ren, Urtasun, and
  Grosse]{gomez_reversible_2017}
Gomez, A.~N., Ren, M., Urtasun, R., and Grosse, R.~B.
\newblock The {Reversible} {Residual} {Network}: {Backpropagation} {Without}
  {Storing} {Activations}.
\newblock \emph{arXiv:1707.04585 [cs]}, July 2017.
\newblock URL \url{http://arxiv.org/abs/1707.04585}.
\newblock arXiv: 1707.04585.

\bibitem[Goodfellow et~al.(2014)Goodfellow, Pouget-Abadie, Mirza, Xu,
  Warde-Farley, Ozair, Courville, and Bengio]{goodfellow_generative_2014}
Goodfellow, I., Pouget-Abadie, J., Mirza, M., Xu, B., Warde-Farley, D., Ozair,
  S., Courville, A., and Bengio, Y.
\newblock Generative {Adversarial} {Nets}.
\newblock In Ghahramani, Z., Welling, M., Cortes, C., Lawrence, N.~D., and
  Weinberger, K.~Q. (eds.), \emph{Advances in {Neural} {Information}
  {Processing} {Systems} 27}, pp.\  2672--2680. Curran Associates, Inc., 2014.
\newblock URL
  \url{http://papers.nips.cc/paper/5423-generative-adversarial-nets.pdf}.

\bibitem[Gretton et~al.(2012)Gretton, Borgwardt, Rasch, Schölkopf, and
  Smola]{gretton_kernel_2012}
Gretton, A., Borgwardt, K.~M., Rasch, M.~J., Schölkopf, B., and Smola, A.
\newblock A {Kernel} {Two}-{Sample} {Test}.
\newblock \emph{Journal of Machine Learning Research}, 13:\penalty0 723--773,
  March 2012.
\newblock ISSN 1533-7928.
\newblock URL \url{http://jmlr.csail.mit.edu/papers/v13/gretton12a.html}.

\bibitem[Grover et~al.(2018)Grover, Dhar, and Ermon]{grover_flow-gan:_2018}
Grover, A., Dhar, M., and Ermon, S.
\newblock Flow-{GAN}: {Combining} {Maximum} {Likelihood} and {Adversarial}
  {Learning} in {Generative} {Models}.
\newblock In \emph{Thirty-{Second} {AAAI} {Conference} on {Artificial}
  {Intelligence}}, 2018.
\newblock URL \url{http://arxiv.org/abs/1705.08868}.
\newblock arXiv: 1705.08868.

\bibitem[Heusel et~al.(2017)Heusel, Ramsauer, Unterthiner, Nessler, and
  Hochreiter]{heusel_gans_2017}
Heusel, M., Ramsauer, H., Unterthiner, T., Nessler, B., and Hochreiter, S.
\newblock {GANs} {Trained} by a {Two} {Time}-{Scale} {Update} {Rule} {Converge}
  to a {Local} {Nash} {Equilibrium}.
\newblock In Guyon, I., Luxburg, U.~V., Bengio, S., Wallach, H., Fergus, R.,
  Vishwanathan, S., and Garnett, R. (eds.), \emph{Advances in {Neural}
  {Information} {Processing} {Systems} 30}, pp.\  6626--6637. Curran
  Associates, Inc., 2017.
\newblock URL
  \url{http://papers.nips.cc/paper/7240-gans-trained-by-a-two-time-scale-update-rule-converge-to-a-local-nash-equilibrium.pdf}.

\bibitem[Isola et~al.(2017)Isola, Zhu, Zhou, and
  Efros]{isola_image--image_2017}
Isola, P., Zhu, J.~Y., Zhou, T., and Efros, A.~A.
\newblock Image-to-{Image} {Translation} with {Conditional} {Adversarial}
  {Networks}.
\newblock In \emph{2017 {IEEE} {Conference} on {Computer} {Vision} and
  {Pattern} {Recognition} ({CVPR})}, pp.\  5967--5976, July 2017.
\newblock \doi{10.1109/CVPR.2017.632}.

\bibitem[Jacobsen et~al.(2018)Jacobsen, Smeulders, and
  Oyallon]{jacobsen_i-revnet:_2018}
Jacobsen, J.-H., Smeulders, A. W.~M., and Oyallon, E.
\newblock i-{RevNet}: {Deep} {Invertible} {Networks}.
\newblock In \emph{International {Conference} on {Learning} {Representations}
  ({ICLR})}, 2018.
\newblock URL \url{https://openreview.net/forum?id=HJsjkMb0Z}.

\bibitem[Karras et~al.(2018)Karras, Aila, Laine, and
  Lehtinen]{karras_progressive_2018}
Karras, T., Aila, T., Laine, S., and Lehtinen, J.
\newblock Progressive {Growing} of {GANs} for {Improved} {Quality},
  {Stability}, and {Variation}.
\newblock In \emph{International {Conference} on {Learning} {Representations}
  ({ICLR})}, 2018.
\newblock URL \url{http://arxiv.org/abs/1710.10196}.
\newblock arXiv: 1710.10196.

\bibitem[Kingma \& Welling(2014)Kingma and Welling]{kingma_auto-encoding_2014}
Kingma, D.~P. and Welling, M.
\newblock Auto-{Encoding} {Variational} {Bayes}.
\newblock In \emph{International {Conference} on {Machine} {Learning}}, 2014.
\newblock URL \url{http://arxiv.org/abs/1312.6114}.
\newblock arXiv: 1312.6114.

\bibitem[Kolouri et~al.(2018)Kolouri, Martin, and
  Rohde]{kolouri_sliced-wasserstein_2018}
Kolouri, S., Martin, C.~E., and Rohde, G.~K.
\newblock Sliced-{Wasserstein} {Autoencoder}: {An} {Embarrassingly} {Simple}
  {Generative} {Model}.
\newblock \emph{arXiv:1804.01947 [cs, stat]}, April 2018.
\newblock URL \url{http://arxiv.org/abs/1804.01947}.
\newblock arXiv: 1804.01947.

\bibitem[Lecun et~al.(1998)Lecun, Bottou, Bengio, and
  Haffner]{lecun_gradient-based_1998}
Lecun, Y., Bottou, L., Bengio, Y., and Haffner, P.
\newblock Gradient-{Based} {Learning} {Applied} to {Document} {Recognition}.
\newblock In \emph{Proceedings of the {IEEE}}, pp.\  2278--2324, 1998.

\bibitem[Li et~al.(2017)Li, Chang, Cheng, Yang, and Poczos]{li_mmd_2017}
Li, C.-L., Chang, W.-C., Cheng, Y., Yang, Y., and Poczos, B.
\newblock {MMD} {GAN}: {Towards} {Deeper} {Understanding} of {Moment}
  {Matching} {Network}.
\newblock In Guyon, I., Luxburg, U.~V., Bengio, S., Wallach, H., Fergus, R.,
  Vishwanathan, S., and Garnett, R. (eds.), \emph{Advances in {Neural}
  {Information} {Processing} {Systems} 30}, pp.\  2203--2213. Curran
  Associates, Inc., 2017.
\newblock URL
  \url{http://papers.nips.cc/paper/6815-mmd-gan-towards-deeper-understanding-of-moment-matching-network.pdf}.

\bibitem[Lim \& Ye(2017)Lim and Ye]{lim_geometric_2017}
Lim, J.~H. and Ye, J.~C.
\newblock Geometric {GAN}.
\newblock \emph{arXiv:1705.02894 [cond-mat, stat]}, May 2017.
\newblock URL \url{http://arxiv.org/abs/1705.02894}.
\newblock arXiv: 1705.02894.

\bibitem[Liu et~al.(2015)Liu, Luo, Wang, and Tang]{liu_deep_2015}
Liu, Z., Luo, P., Wang, X., and Tang, X.
\newblock Deep {Learning} {Face} {Attributes} in the {Wild}.
\newblock In \emph{Proceedings of {International} {Conference} on {Computer}
  {Vision} ({ICCV})}, December 2015.

\bibitem[Lucic et~al.(2017)Lucic, Kurach, Michalski, Gelly, and
  Bousquet]{lucic_are_2017}
Lucic, M., Kurach, K., Michalski, M., Gelly, S., and Bousquet, O.
\newblock Are {GANs} {Created} {Equal}? {A} {Large}-{Scale} {Study}.
\newblock \emph{arXiv:1711.10337 [cs, stat]}, November 2017.
\newblock URL \url{http://arxiv.org/abs/1711.10337}.
\newblock arXiv: 1711.10337.

\bibitem[Makhzani et~al.(2016)Makhzani, Shlens, Jaitly, Goodfellow, and
  Frey]{makhzani_adversarial_2016}
Makhzani, A., Shlens, J., Jaitly, N., Goodfellow, I., and Frey, B.
\newblock Adversarial {Autoencoders}.
\newblock In \emph{International {Conference} on {Learning} {Representations}
  ({ICLR})}, 2016.
\newblock URL \url{http://arxiv.org/abs/1511.05644}.
\newblock arXiv: 1511.05644.

\bibitem[Miyato et~al.(2018)Miyato, Kataoka, Koyama, and
  Yoshida]{miyato_spectral_2018}
Miyato, T., Kataoka, T., Koyama, M., and Yoshida, Y.
\newblock Spectral {Normalization} for {Generative} {Adversarial} {Networks}.
\newblock In \emph{International {Conference} on {Learning} {Representations}
  ({ICLR})}, 2018.
\newblock URL \url{http://arxiv.org/abs/1802.05957}.
\newblock arXiv: 1802.05957.

\bibitem[Peyré \& Cuturi(2018)Peyré and Cuturi]{peyre_computational_2018}
Peyré, G. and Cuturi, M.
\newblock Computational {Optimal} {Transport}.
\newblock \emph{arXiv:1803.00567 [stat]}, March 2018.
\newblock URL \url{http://arxiv.org/abs/1803.00567}.
\newblock arXiv: 1803.00567.

\bibitem[Rubenstein et~al.(2018)Rubenstein, Schoelkopf, and
  Tolstikhin]{rubenstein_latent_2018}
Rubenstein, P.~K., Schoelkopf, B., and Tolstikhin, I.
\newblock On the {Latent} {Space} of {Wasserstein} {Auto}-{Encoders}.
\newblock In \emph{International {Conference} on {Learning} {Representations}
  ({ICLR}) {Workshop}}, 2018.
\newblock URL \url{http://arxiv.org/abs/1802.03761}.
\newblock arXiv: 1802.03761.

\bibitem[Salimans et~al.(2018)Salimans, Zhang, Radford, and
  Metaxas]{salimans_improving_2018}
Salimans, T., Zhang, H., Radford, A., and Metaxas, D.
\newblock Improving {GANs} {Using} {Optimal} {Transport}.
\newblock In \emph{International {Conference} on {Learning} {Representations}
  ({ICLR})}, 2018.
\newblock URL \url{https://openreview.net/forum?id=rkQkBnJAb}.

\bibitem[Shang et~al.(2016)Shang, Sohn, Almeida, and
  Lee]{shang_understanding_2016}
Shang, W., Sohn, K., Almeida, D., and Lee, H.
\newblock Understanding and improving convolutional neural networks via
  concatenated rectified linear units.
\newblock In \emph{International {Conference} on {Machine} {Learning}}, pp.\
  2217--2225, 2016.

\bibitem[Tolstikhin et~al.(2018)Tolstikhin, Bousquet, Gelly, and
  Schoelkopf]{tolstikhin_wasserstein_2018}
Tolstikhin, I., Bousquet, O., Gelly, S., and Schoelkopf, B.
\newblock Wasserstein {Auto}-{Encoders}.
\newblock In \emph{International {Conference} on {Learning} {Representations}
  ({ICLR})}, 2018.
\newblock URL \url{http://arxiv.org/abs/1711.01558}.
\newblock arXiv: 1711.01558.

\bibitem[Ulyanov et~al.(2017)Ulyanov, Vedaldi, and
  Lempitsky]{ulyanov_adversarial_2017}
Ulyanov, D., Vedaldi, A., and Lempitsky, V.
\newblock Adversarial {Generator}-{Encoder} {Networks}.
\newblock \emph{arXiv preprint arXiv:1704.02304}, 2017.

\bibitem[Zhang et~al.(2018)Zhang, Goodfellow, Metaxas, and
  Odena]{zhang_self-attention_2018}
Zhang, H., Goodfellow, I., Metaxas, D., and Odena, A.
\newblock Self-{Attention} {Generative} {Adversarial} {Networks}.
\newblock \emph{arXiv:1805.08318 [cs, stat]}, May 2018.
\newblock URL \url{http://arxiv.org/abs/1805.08318}.
\newblock arXiv: 1805.08318.

\end{thebibliography}
\bibliographystyle{icml2018}

\end{document}